\documentclass {article}

\usepackage[latin1]{inputenc}
\usepackage{amsmath}
\usepackage{amsfonts}
\usepackage{amssymb}
\usepackage{graphicx}
\usepackage{bm}
\usepackage{subcaption}
\usepackage{xcolor}
\usepackage{algorithm}
\usepackage{algorithmic}

\usepackage[accepted]{icml2020}

    \newcommand{\VT}[1]{\ensuremath{{V_{T#1}}}}

    \newcommand{\bfx}  	{\mathbf{x}}
    \newcommand{\bfy}  	{\mathbf{y}}
    \newcommand{\bfz}  	{\mathbf{z}}

    \newbox\sectsavebox
    \setbox\sectsavebox=\hbox{\boldmath\VT{xyz}}
    
\icmltitlerunning{Outlier Detection through Null Space Analysis of Neural Networks}

\begin{document}

    \twocolumn[
    \icmltitle{Outlier Detection through Null Space Analysis of Neural Networks}
    
    \begin{icmlauthorlist}
        \icmlauthor{Matthew Cook}{ece}
        \icmlauthor{Alina Zare}{ece}
        \icmlauthor{Paul Gader}{cise}
    \end{icmlauthorlist}
    
    \icmlaffiliation{ece}{Department of Electrical \& Computer Engineering University of Florida, Gainesville Fl, USA 32611}
    \icmlaffiliation{cise}{Department of Computer \& Information Science \& Engineering University of Florida, Gainesville Fl, USA 32611}
    
    \icmlcorrespondingauthor{Matthew Cook}{matthew.cook@ufl.edu}
    
    \icmlkeywords{Artificial Neural Networks, Machine Learning, Null Space, Outlier Detection}
    
    \vskip 0.3in
    ]

    \printAffiliationsAndNotice{}
    
    \begin{abstract}
        Many machine learning classification systems lack competency awareness.  Specifically, many systems lack the ability to identify when outliers (e.g., samples that are distinct from and not represented in the training data distribution) are being presented to the system.  The ability to detect outliers is of practical significance since it can help the system behave in an reasonable way when encountering unexpected data. In prior work, outlier detection is commonly carried out in a processing pipeline that is distinct from the classification model. Thus, for a complete system that incorporates outlier detection and classification, two models must be trained, increasing the overall complexity of the approach. In this paper we use the concept of the null space to integrate an outlier detection method directly into a neural network used for classification. Our method, called Null Space Analysis (NuSA) of neural networks, works by computing and controlling the magnitude of the null space projection as data is passed through a network. Using these projections, we can then calculate a score that can differentiate between normal and abnormal data. Results are shown that indicate networks trained with NuSA retain their classification performance while also being able to detect outliers at rates similar to commonly used outlier detection algorithms.
    \end{abstract}
  
    \section{Introduction}
        \label{Intr}
        
        Artificial neural network (ANN) and deep learning-based approaches are increasingly being incorporated into real-world applications and systems.  Unlike curated benchmark datasets, outliers and unexpected data is commonly encountered in systems deployed for application.  Thus, the ability to accurately identify outliers to ensure reliability of ANN-dependent systems is essential.  Due to this, outlier detection has been a key aspect of machine learning for some time \cite{MCD, LOF, VAE}.  In standard outlier detection, the goal is to identify inputs that are distinct from the training data distribution \cite{OutlierDet}. These distinctions can be subtle, as illustrated by the work in adversarial examples \cite{SzegedyIntriguingProps, FawziRobustnessAdversarial}, and also can be samples from an unknown class (e.g., presenting an image of a dog to an ANN trained to differentiate between species of cats). Without mechanisms to detect outliers, ANNs will classify every sample (sometimes with high confidence \cite{MoosaviDeepFool}) to a class found in the training data.  In this paper, we focus on the outlier detection paradigm of unknown classes and propose a method for ANNs to identify when data does not match its trained purpose. 
        
        A variety of outlier detection methods have been developed in the literature.  A common outlier detection method is to use distances to nearby points \cite{LOF, HBOS}. An example of this is to use the distance to the $k$ nearest neighbors as an outlier score \cite{KNN}. Another example is the Angle-Based Outlier Detection \cite{ABOD} which uses both the distance and angle between points to identify outliers. Another class of outlier detection algorithms are one-class classifiers. A popular example of these algorithms is the One-Class Support Vector Machine \cite{OCSVM}. In this case the Support Vector Machine (SVM) is used to estimate a hyperplane that encompasses the training data and outliers can be detected based on their distance from the hyperplane. The major and minor principal components of known ``good'' data to generate a hyperplane for comparison has also been used \cite{PCA}. The Isolation Forest \cite{IForest} is another method that uses a random forest to find outliers. Methods that identify outliers by having a large reconstruction error using a model fit with the training data have also been used.  Recently, the most popular examples of these methods are auto-encoders \cite{AutoEnc} or variational auto-encoders \cite{VAE}. However, all of these methods are capable of only one thing, outlier detection. We propose a method using Null Space Analysis (NuSA) that is capable of encoding outlier detection directly into an ANN, thereby increasing the competency awareness of the ANN.
        
        In our NuSA approach, we leverage the null space associated with the weight matrix of a layer in the ANN to perform outlier detection concurrently with classification using the same network. As a refresher, the null space of a matrix is defined as:
        \begin{equation}
            \mathcal{N}(\bm{A}) = \left\{\bm{z}\in\mathcal{R}^n\left|\bm{Az}=\bm{0}
            \right.\right\},
            \label{Nspace}
        \end{equation}
        where $\bm{A}$ is a linear mapping. In other words, the null space of matrix, $\bm{A}$, defines the region of the input space that maps to 
        zero. The motivation to leverage the null space is related to the study of adversarial samples such as those shown in \cite{NguyenSeminal} and to experiences in handwritten word recognition in the 1990s \cite{ChiangFuzzNeur, FuzzNeurComp}. The NuSA approach is a partial, but important,  solution to the problem of competency awareness of ANNs; it is unlikely that there is one method alone that can alleviate this problem. Outlier samples derived from the null space of a weight matrix of a network can theoretically have infinite magnitude and added to non-outlier samples with no deviation in the ANN output. This statement will be made more precise in the next section.

    \section{Null Space Analysis of ANNs}
        \label{sec:NuSANN}
       
        Each layer of an ANN is the projection of a sample by a weight matrix followed by the application of an activation function.  This sample is either the input data point at the first layer or the output of a previous layer for any hidden layers. Every weight matrix has an associated null space, although some may be empty.  However, any weight matrix that projects into a lower dimensional space (i.e., input dimensionality is larger than the output/subsequent layer dimensionality) has a non-empty null space.  The overwhelming majority of all commonly used deep learning architectures consist of several subsequent layers that project into lower dimensional spaces and, thus, have an associated series of non-empty null spaces.  Any sample with a non-zero projection into any null space in the network, cannot be distinguished by the network from those samples without that null space component.  
        
        For clarity, the null space concept is first illustrated using a simple one-layer ANN with $K$ inputs and $M$ outputs and $K > M$. It is then illustrated for  for multi-layer networks. Assume $X = \left\{\bfx_1,\bfx_2,\dots,\bfx_N\right\}$ is a collection of samples drawn from the joint distribution of the classes of interest, that is,  $\bfx_n\sim p_{\mathcal{D}}$ and that a network, $f_X$ has been trained to approximate a function $f$ with $\bfy_n = f \left(\bfx_n \right)$. In the one-layer case, $\bfy_n = W\bfx_n$. Since $K > M$, there is a non-trivial null space $\mathcal{N}(W)$ of dimension $K-M$.   If  $\bfz \in \mathcal{N}(W)$, then $W\bfz=\mathbf{0}$  so 

        \begin{equation}
            \begin{aligned}
                \forall{\lambda > 0}\hspace{12pt} W(\bfx_n + \lambda\bfz) =& \\
                W\bfx_n+\lambda& W\bfz = W\bfx_n = \bfy_n.
            \end{aligned}
            \label{Eq:NullSpaceProblem}
        \end{equation}
        \noindent which implies that there are infinitely many possibilities for mapping unknown inputs to apparently meaningful outputs. In a multi-layer network, there are many layers of matrix-vector multiplications which can be expressed as
        
        \begin{equation*}
            \begin{aligned}
                \bfx_{n,1} &= \sigma_1(W_1\bfx_n) \rightarrow \bfx_{n,2} \\
                \bfx_{n,2} &= \sigma_2(W_2\bfx_{n,1}) \rightarrow \cdots \bfx_{n,N_h} \\
                \bfx_{n,N_h} &=  \sigma_{N_H}(W_{N_H}\bfx_{n,N_H-1}) 
            \end{aligned}
            \label{Eq:NetFormula}
        \end{equation*}
        \noindent where $\bfx$ is an input,   $N_H$ is the number of hidden layers, and $\sigma_h, h=1,2,\dots,N_H$ are nonlinear functions.  More succinctly
        
        \begin{equation}
            \begin{aligned}
                f_X&(\bfx_n) = \\
                &\sigma_{N_h}(W_{N_h}\sigma_{N_h-1}(W_{N_h-1}\sigma( \cdots\sigma_1(W_1\bfx)\cdots)))
            \end{aligned}
            \label{Eq:FNet}
        \end{equation}
        
        %\noindent If $\bfx,\bfz$ are input samples, and $\bfx_h-\bfz_h \in \mathcal{N}(W_{h+1})$ at layer $h+1$ in the network, then $W_{h+1}\bfx_h = W_{h+1}\bfz_h$ so $\bfx_m = \bfz_m$ for $m = h+1,\dots,N_h$.
        
        Consider a non-outlier sample $\bm{x} \sim p_{\mathcal{D}}$. Let $\bfz \in \mathbb{R}_K$ and $\bfz_h =  \sigma_{h-1}(W_{h-1}\sigma( \cdots\sigma_1(W_1\bfx)))$ for any $h = 1,2,\dots,N_h$. If $\bfz_h \in \mathcal{N}(W_h)$ then $W_h\bfz_h = \mathbf{0}$ so $f_X(\bfx+\bfz) = f_X(\bfx)$.  Therefore, any input sample, $\bfz$,  that is the inverse image of a sample, $\bfz_h$ of the null space of any of the weight matrices, $W_h$, can be added to a legitimate input, $\bfx$ and the output will not change.  This is one source of ``adversarial'' examples that can cause outliers that are nothing like any of the true samples to have high outputs for at least one class.
        
\begin{figure}[hbt]
            \centering
            \includegraphics[width=0.32\linewidth,trim={1.25cm 1.25cm 1.25cm 1.25cm},clip]{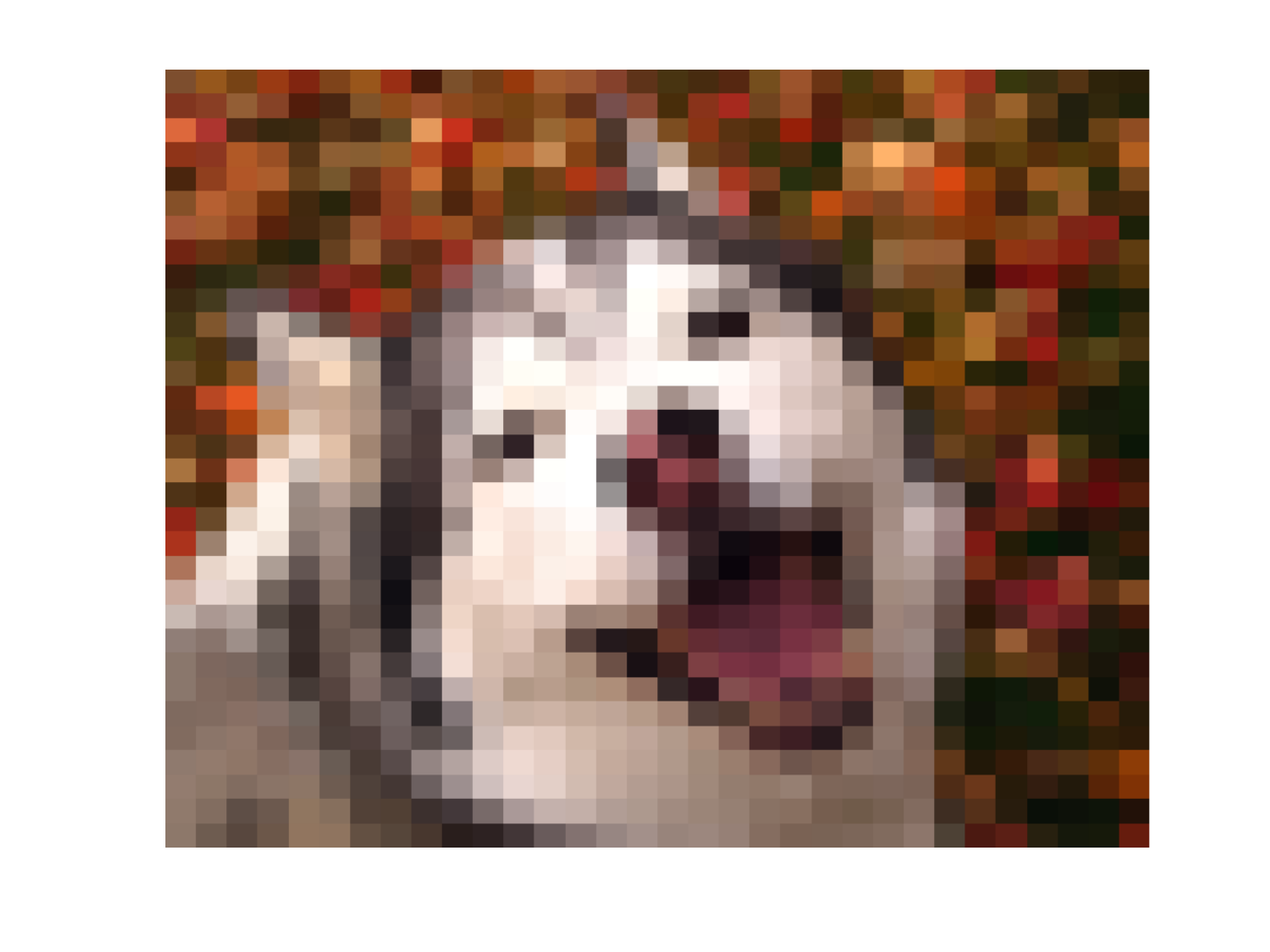}
            \includegraphics[width=0.32\linewidth,trim={1.25cm 1.25cm 1.25cm 1.25cm},clip]{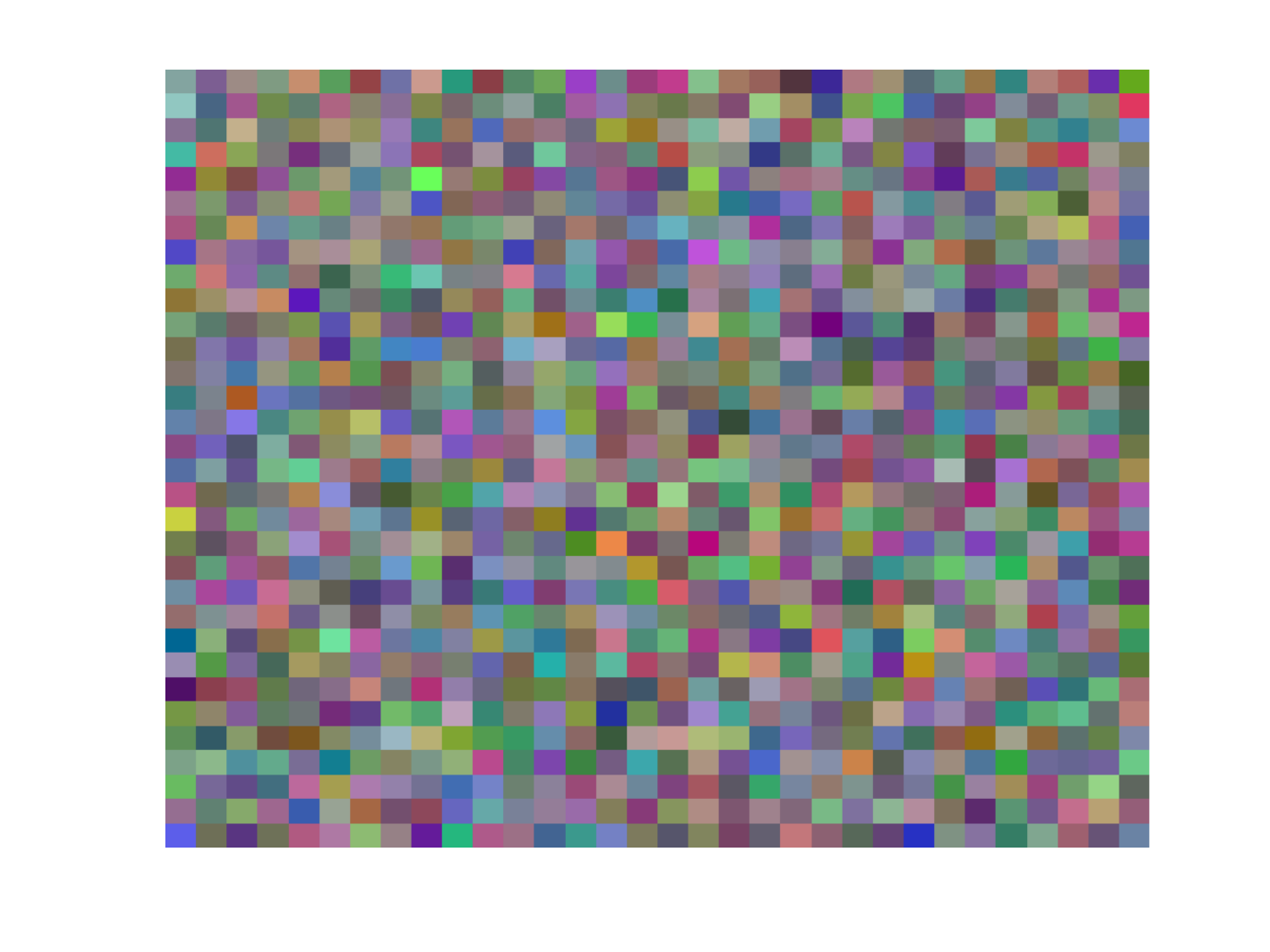}
            \includegraphics[width=0.32\linewidth,trim={1.25cm 1.25cm 1.25cm 1.25cm},clip]{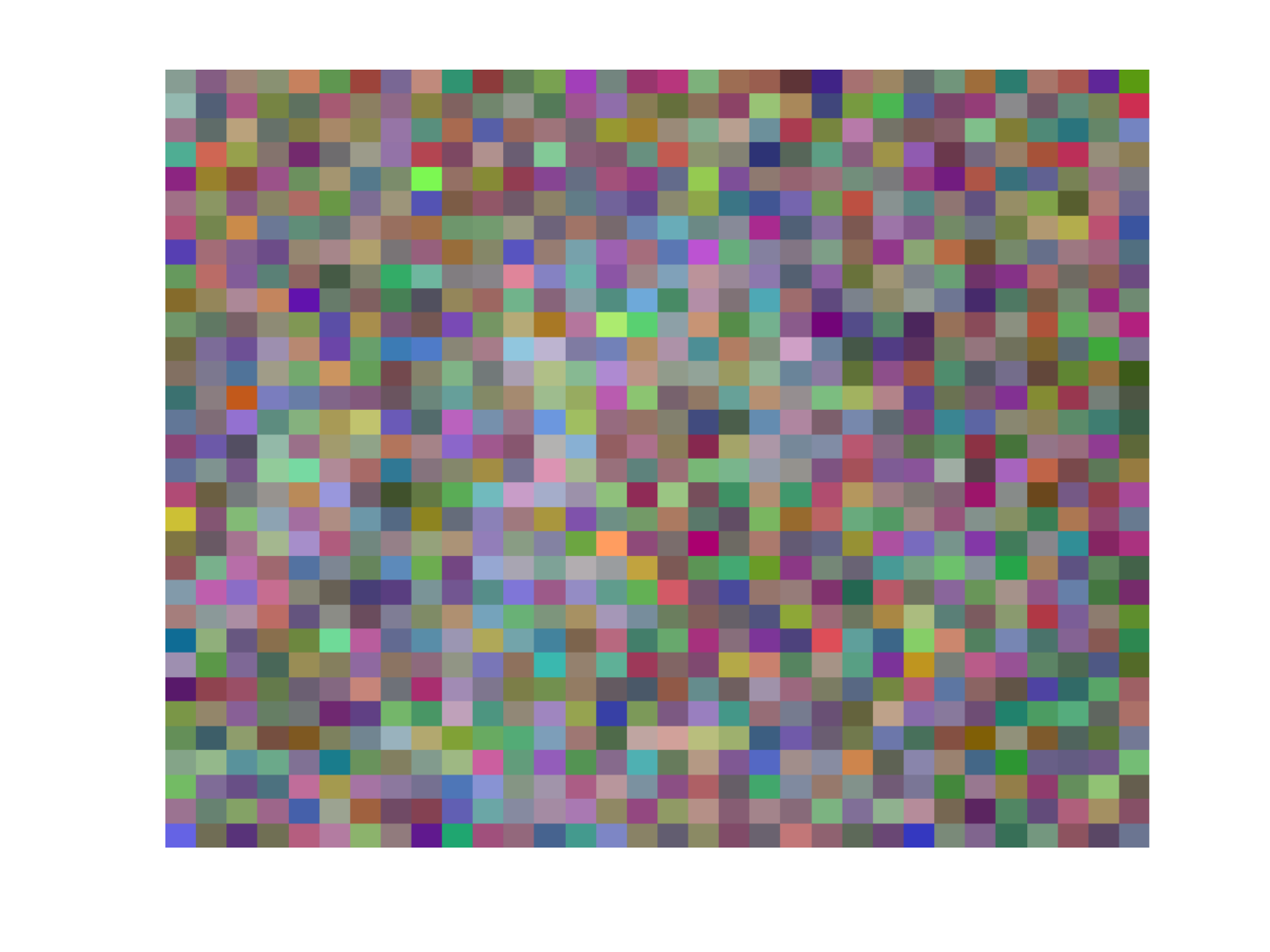}
            \caption{Left: Original image. Center: Additive null space noise. Right: Final image, indistinguishable from original image according to the network the noise in the center column is sampled from.}
            \label{fig:NSnoise}
        \end{figure}

        The description above outlines an interesting and unique  class of adversarial samples for a network.  Some adversarial samples are defined via stability,  a small change in an input sample (e.g., imperceptible to a human) can produce a large change in output.  The NuSA approach is focused on the opposite problem, i.e., large changes in an input sample can produce a small (or, no) changes in output.  A human would easily disregard this heavily corrupted sample as an outlier but, as pointed out in \cite{ChiangFuzzNeur, FuzzNeurComp, NguyenSeminal},  the network would not be able to distinguish the sample from the valid sample.  An example of this is shown in Figure \ref{fig:NSnoise}. 
        
        In our approach, we aim to leverage the null space to identify these outliers.  The NuSA approach is to maximize the projection of any training data samples into the null space of layers in a network. Then, during tests, the magnitude of the projection onto to the null space can be monitored and any sample with a large null space projection can be flagged as an outlier. The idea is to push everything into the null except for the classes that exist in the training sets. This is difficult for the network since by maximizing the null space projection most of the data is lost. 
        
        In order to accomplish this, we define the Null Space Analysis (NuSA) Term which computes the magnitude of the projecting onto the span of the column space as in \eqref{eqn:NuSA}: 
        \begin{equation}
            \mathcal{N}_u\mathcal{SA} = \frac{\|\mathcal{P}(\bm{W})\bm{x}\|}{\|\bm{x}\|},
            \label{eqn:NuSA}
        \end{equation}
        where $\bm{x}$ and $\bm{W}$ are respectively the input sample and weight matrix 
        of the network layer being considered and $\mathcal{P}(.)$ is the 
        projection matrix defined as \eqref{eqn:proj}:
        \begin{equation}
            \mathcal{P}(\bm{W}) = \bm{W}^T(\bm{WW}^T)^{-1}\bm{W}.
            \label{eqn:proj}
        \end{equation}
        This term can be interpreted as comparing the length of the original input sample to the length of the input sample once projected onto the column space. The column space of the matrix is used (instead of the null space directly) as it is much easier to compute and projections onto the column space are inversely related to null space projections. The column space of a matrix is said to be all possible linear combinations of the columns of the matrix. Thus no calculations are needed to find the column space.
        
        In practice, NuSA is calculated in two steps. First the QR decomposition of the weight matrix is calculated to find an orthogonal column space basis of the weight matrix. Normalizing the columns of the column space basis leaves an orthonormal basis meaning Equation \ref{eqn:proj} can be simplified to 
        \begin{equation}
            \mathcal{P}(\bm{W}) = \bm{\mathcal{C}}(\bm{W})^T\bm{\mathcal{C}}(\bm{W}),
        \end{equation}
        where the function $\bm{\mathcal{C}}(\bm{W})$ represents the column space basis of the matrix $\bm{W}$. This step makes calculating the NuSA statistic much easier.
        
    \section{Outlier Detection with Null Space Analysis}
        \label{sec:ODwNuSA}
        
        Now that we have defined a function that computes the magnitude of the null space projection we now apply this to a ANN to find outliers. We can encode the outlier detection into a network by directly incorporating the NuSA calculation into the loss function used for training. In order to incorporate NuSA into network training, \eqref{eqn:NuSAFull} is added as a term in the loss function.  This term sums the evaluation of \eqref{eqn:NuSA} over each fully connected layer in the network, 
        \begin{equation}
            \mathcal{N}_u\mathcal{SA} = \lambda\sum\limits_{l\in L}\frac{\|\mathcal{P}(\bm{W}_l)\bm{x}_l\|}{\|\bm{x}_l\|}
            \label{eqn:NuSAFull}
        \end{equation}
        where $L$ is the set of layers that of fully connected layers, $x_l$ the input sample of the $l$\textsuperscript{th} layer of the network, and $\bm{W}_l$  is the weight for the $l$\textsuperscript{th} layer of the network. The parameter $\lambda$ controls the tradeoff between minimizing the standard loss function or the NuSA term. Therefore the full loss function used during training can be seen in Equation \ref{eqn:FullObj}.
        \begin{equation}
            \mathcal{L}(\bm{\theta}, \bm{x}) + \lambda\sum\limits_{l\in L}\frac{\|\mathcal{P}(\bm{W}_l)\bm{x}_l\|}{\|\bm{x}_l\|}
            \label{eqn:FullObj}
        \end{equation}
        In this equation we have the the standard loss function $\mathcal{L}(\bm{\theta}, \bm{x})$, combined with our new NuSA term. The network is trained using backpropagation once the loss function has been changed. However, there is a new step during testing shown in Algorithm \ref{alg:Test}. During testing the NuSA score is computed alongside the normal network output. A threshold can then be applied to the NuSA score and values too low can be eliminated as outliers.
        
        \begin{algorithm}
            \caption{Psuedocode for NuSA testing procedure.}
            \label{alg:Test}
            \begin{algorithmic}
                \FOR {i $<$ N samples}
                    \STATE Compute forward pass to get output for sample i
                    \STATE Compute NuSA for sample i
                    \IF {NuSA $>$ threshold}
                        \STATE Set output as index of max network output
                    \ELSE
                        \STATE Declare sample i as outlier
                    \ENDIF
                \ENDFOR
                \STATE \textbf{Return: }Outlier indicator and outlier class labels
            \end{algorithmic}
        \end{algorithm}

    \section{Experiments}
        \label{sec:Exp}
        
        To evaluate the NuSA approach, we studied the impact the inclusion of the NuSA term has on classification performance and its ability to detect outliers using the CIFAR10 dataset \cite{CifarDatasets}. The CIFAR10 dataset consists of many $32\times 32$ images. At this time, NuSA has only been applied to fully connected layers of ANNs. Convolutional ANNs outperform networks made of only fully connected layers in most image datasets. Therefore we use the WideResNet architecture \cite{WideResNet} to generate features that we use to train our basic fully connected networks. In particular, the output layer from WideResNet is removed and the features use for our experiments are the new output (essentially the input to the original output layer). WideResNet claims to have achieved state-of-the-art performance on the CIFAR10 dataset so the features generated from this network can be assumed to be of high quality. WideResNet was trained on $95\%$ of the CIFAR10 training data (47,500 samples) with the remaining $5\%$ (2,500 samples) used as validation, the splits were made randomly.
        
\begin{figure}[!htb]
            \centering
            \includegraphics[width=.75\linewidth]{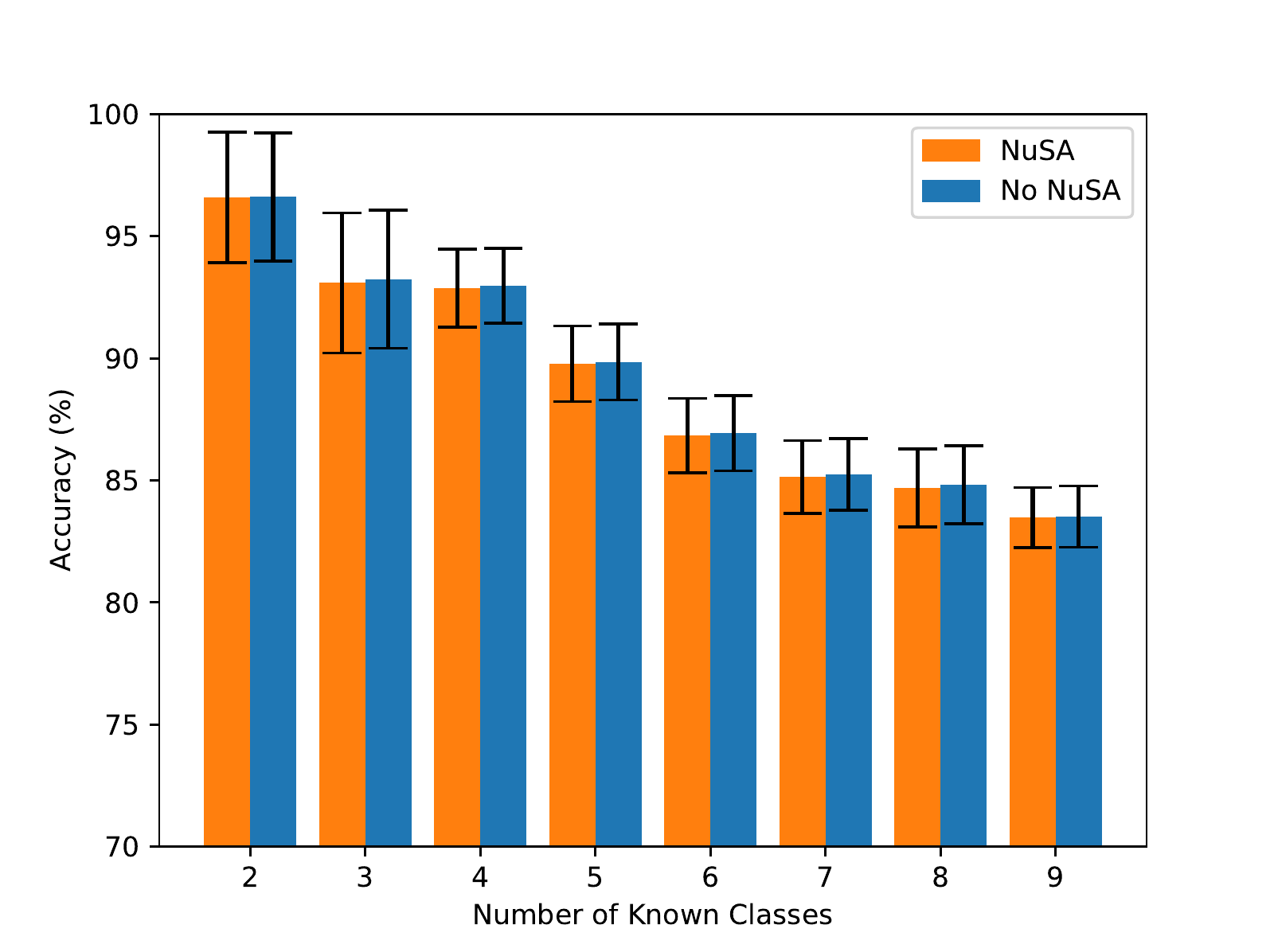}
            \caption{Comparison of classification performance with and without NuSA. One standard deviation is shown by the error bars. Results shown are combined results (average and standard deviation) for the first twenty runs of the models (as returned by the Python function \texttt{itertools.combinations}, except for the case with nine known classes as there were only ten possible combinations. All runs began with a randomly initiated network. Both networks were trained with a batch size of 25, learning rate of 0.01, and a $\lambda$ value for NuSA of 0.1.}
            \label{fig:ClsCmp}
        \end{figure}

        In our experiments, we use a simple network that consists of one hidden layer, 64 inputs with 32 outputs. The number of outputs depends on the test which varies from two to nine. The sigmoid activation function is used along with the Adam optimizer \cite{kingma2014adam}. For each experiment, we use a subset of the CIFAR10 classes to serve as ``known'' classes which we train the classifier against. The remaining classes are used only to serve as outliers at test time. The superset of training data that contains all ten classes is held fixed for all experiments, and is the same as were used for training/validation when WideResNet was trained. The test set is the same for every outlier detection test that follows (outliers are not allowed in the accuracy tests). Every possible combination of known and unknown classes was tested but only a representative subset are shown in the paper for compactness (since there are 1012 number of possible known/unknown class combinations).  Figure \ref{fig:ClsCmp} shows the performance with and without NuSA on CIFAR10 on this network. The goal of this experiment is to verify that the NuSA term does not impact classification performance when included and not to improve or evaluate classification accuracy on CIFAR10 relative to the state of the art in the literature. We can see from the figure that NuSA  has little to no impact on the predictive capability of the trained classifier. This is shown by the average accuracy being nearly identical and heavily overlapping standard deviation bars.
            
        With the effects of NuSA on accuracy shown to be minimal we now move to outlier rejection. For these tests we will be comparing the ability to detect outliers between our simple network with NuSA and several algorithms included in the Python Outlier Detection (PyOD) Toolbox \cite{PyOD}. We will score the algorithms based on how well each method can identify the unknown classes.
        
\begin{figure}[!htb]
            \centering
            \begin{subfigure}[t]{.49\linewidth}
                \centering
                \includegraphics[width=\linewidth]{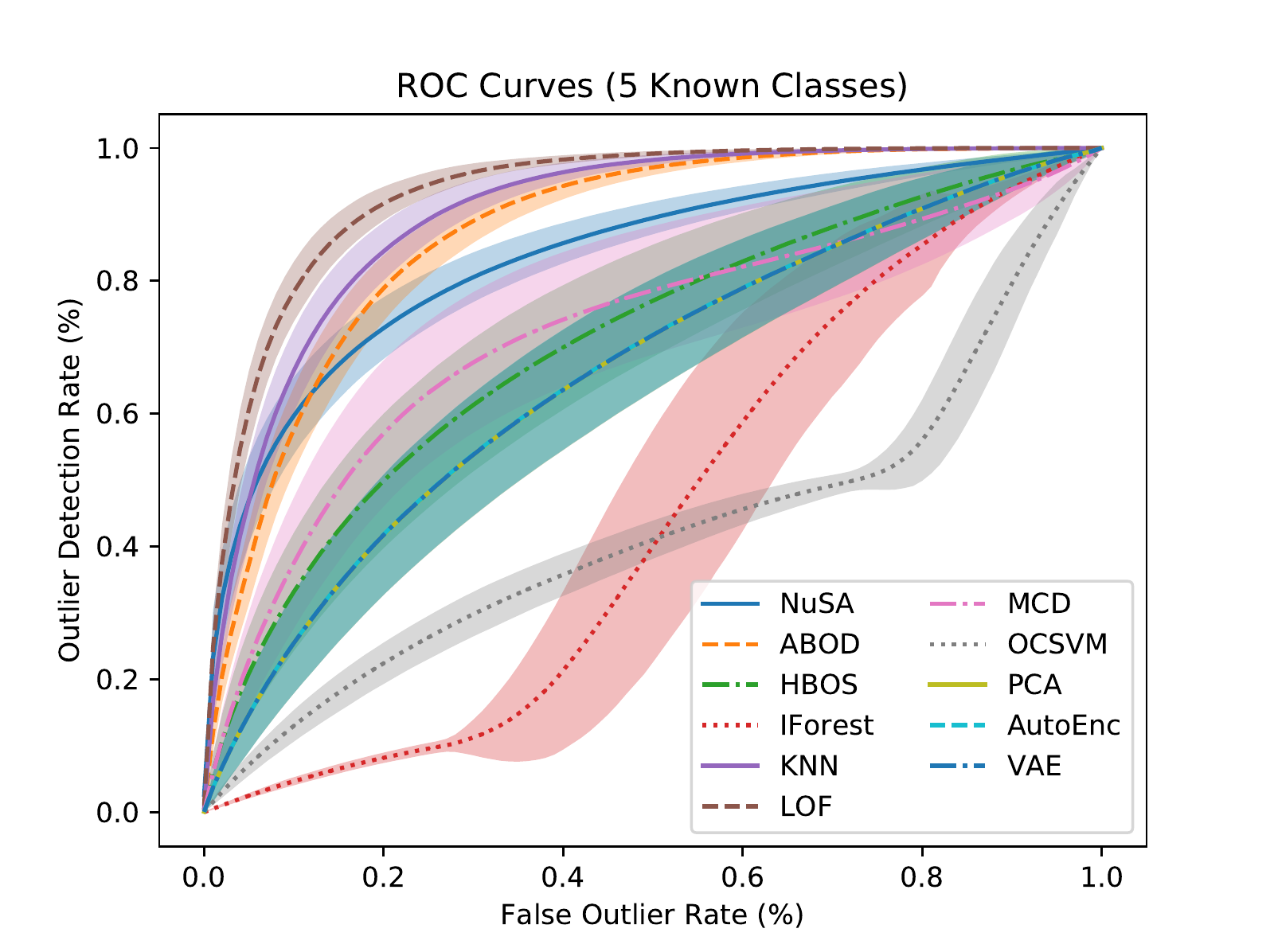}
                \caption{}
                \label{fig:ResROC}
            \end{subfigure}
            \begin{subfigure}[t]{.49\linewidth}
                \centering
                \includegraphics[width=\linewidth]{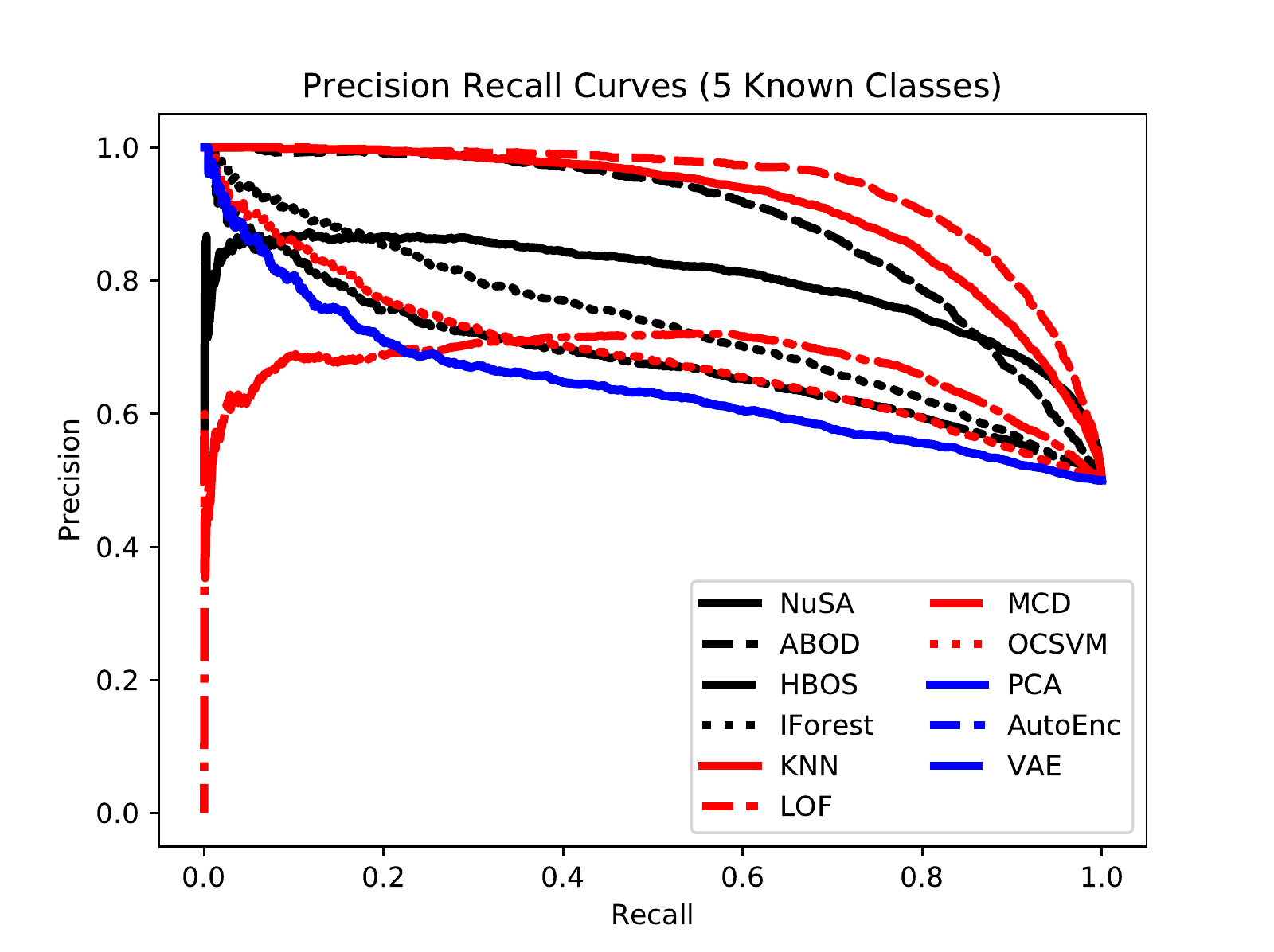}
                \caption{}
                \label{fig:ResPP}
            \end{subfigure}
            \caption{(a) Average ROC curves for each method tested. (b) Precision recall curves for each method tested. Both results are from the case where there are five known classes, therefore each plot in this figure is the average of 252 combinations. }
            \label{fig:ResPlts}
        \end{figure}

        Figure \ref{fig:ResPlts} shows the average ROC and precision recall curves respectively (computed by averaging precision over fixed x-values) for NuSA and several other algorithms. Both of these plots were made use the results from the tests with five known classes. This set of tests has the most individual tests, at 252, and is also the only configuration that is balanced between known and unknown classes. In these results we can see that the simple network with NuSA outperforms several of the dedicated outlier detection algorithms. However, the performance of NuSA is not as good as the performance of Angle Based Outlier Detection \cite{ABOD}, K-Nearest Neighbors Outlier Detection \cite{KNN}, or the Local Outlier Factor \cite{LOF}. Again, an advantage of NuSA is that it is incorporated into the classification ANN.  However, a single strategy/approach for competancy awareness (e.g., NuSA alone) is unlikely to be sufficient in application. 
        
    \section{Conclusion}
        \label{sec:Conc}
        
        In this paper we have presented a new method for detecting outliers during testing. Specifically, NuSA is able to detect outliers without the need to train an additional model exclusively for outlier detection. NuSA is incorporated directly into the ANN used for classification. This has the advantage of only needing to train and run one model at test time as the classification network is now capable of doing both classification and outlier detection simultaneously. While the outlier detection performance of the NuSA network does not quite stack up with state-of-the-art outlier detection algorithms it is an important step towards true competency detection as it provides ANNs the ability to find outliers internally.
    
    \clearpage
    
    \appendix
    
    \section{Additional Results}
        \label{sec:ARes}
        
\begin{figure*}[!htb]
            \centering
            \begin{subfigure}[t]{.24\linewidth}
                \centering
                \includegraphics[width=\textwidth]{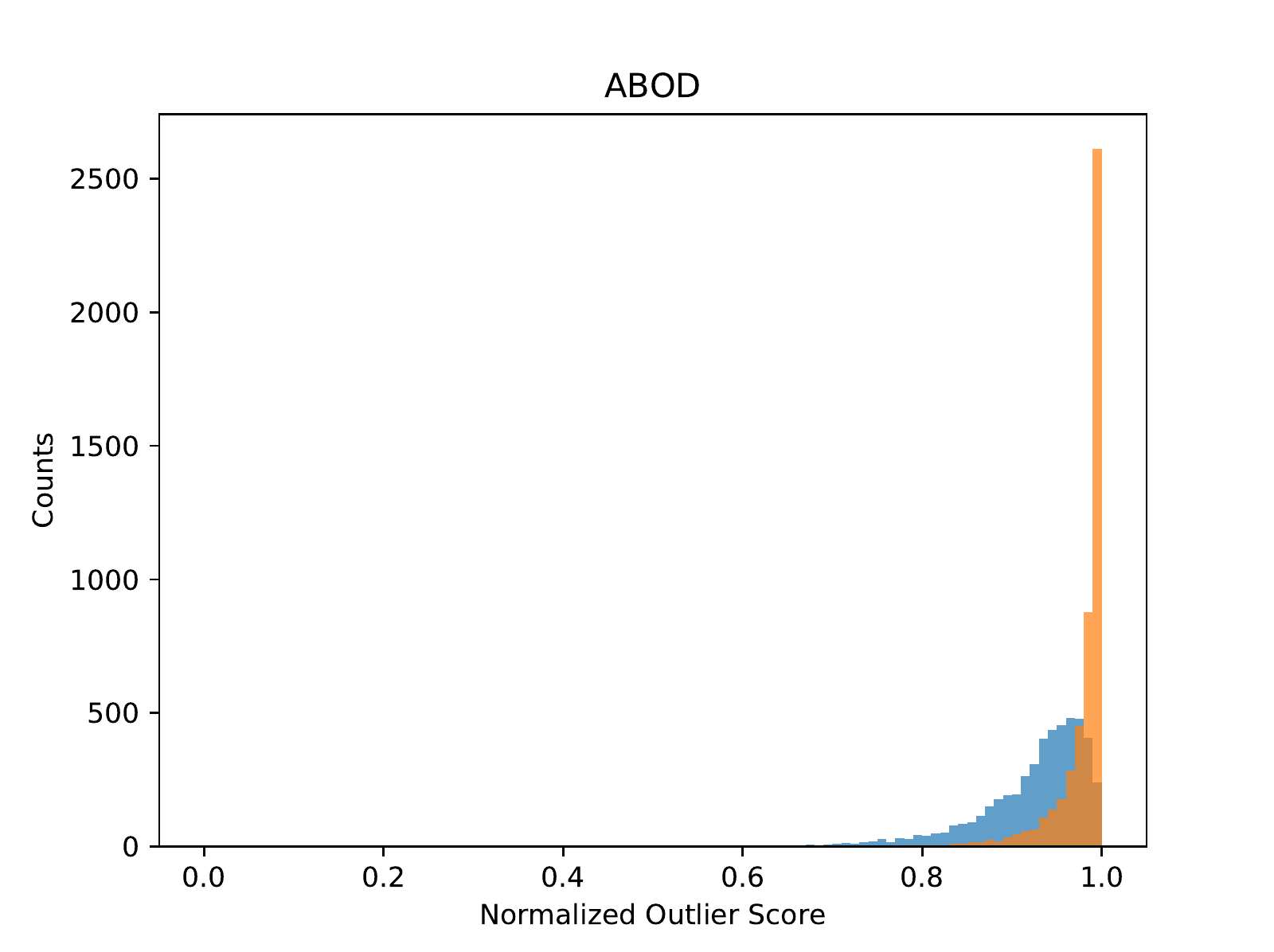}
                \caption{ABOD \cite{ABOD}}
                \label{fig:ABOD_Hist}
            \end{subfigure}
            \begin{subfigure}[t]{.24\linewidth}
                \centering
                \includegraphics[width=\textwidth]{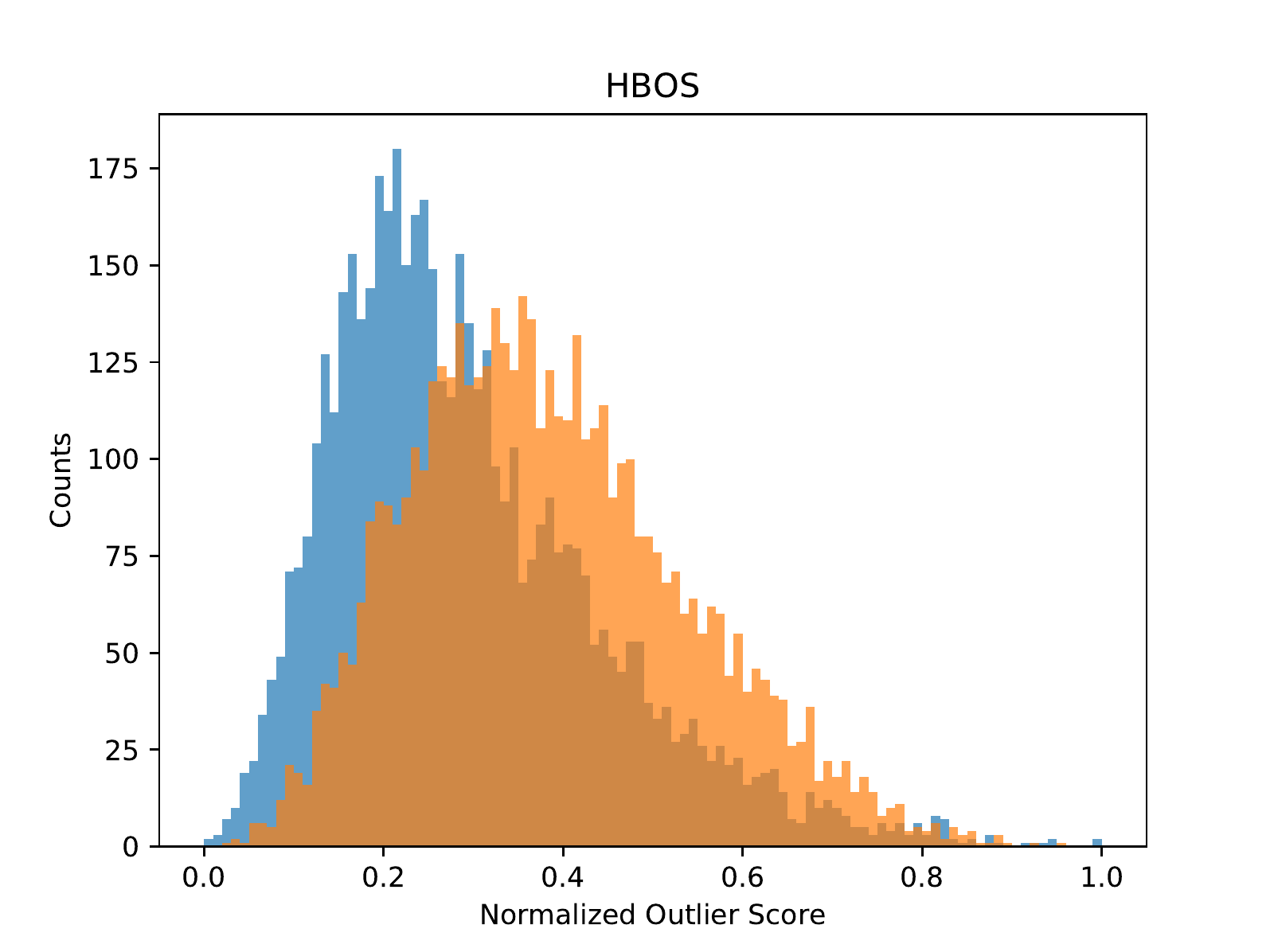}
                \caption{HBOS \cite{HBOS}}
                \label{fig:HBOS_Hist}
            \end{subfigure}
            \begin{subfigure}[t]{.24\linewidth}
                \centering
                \includegraphics[width=\textwidth]{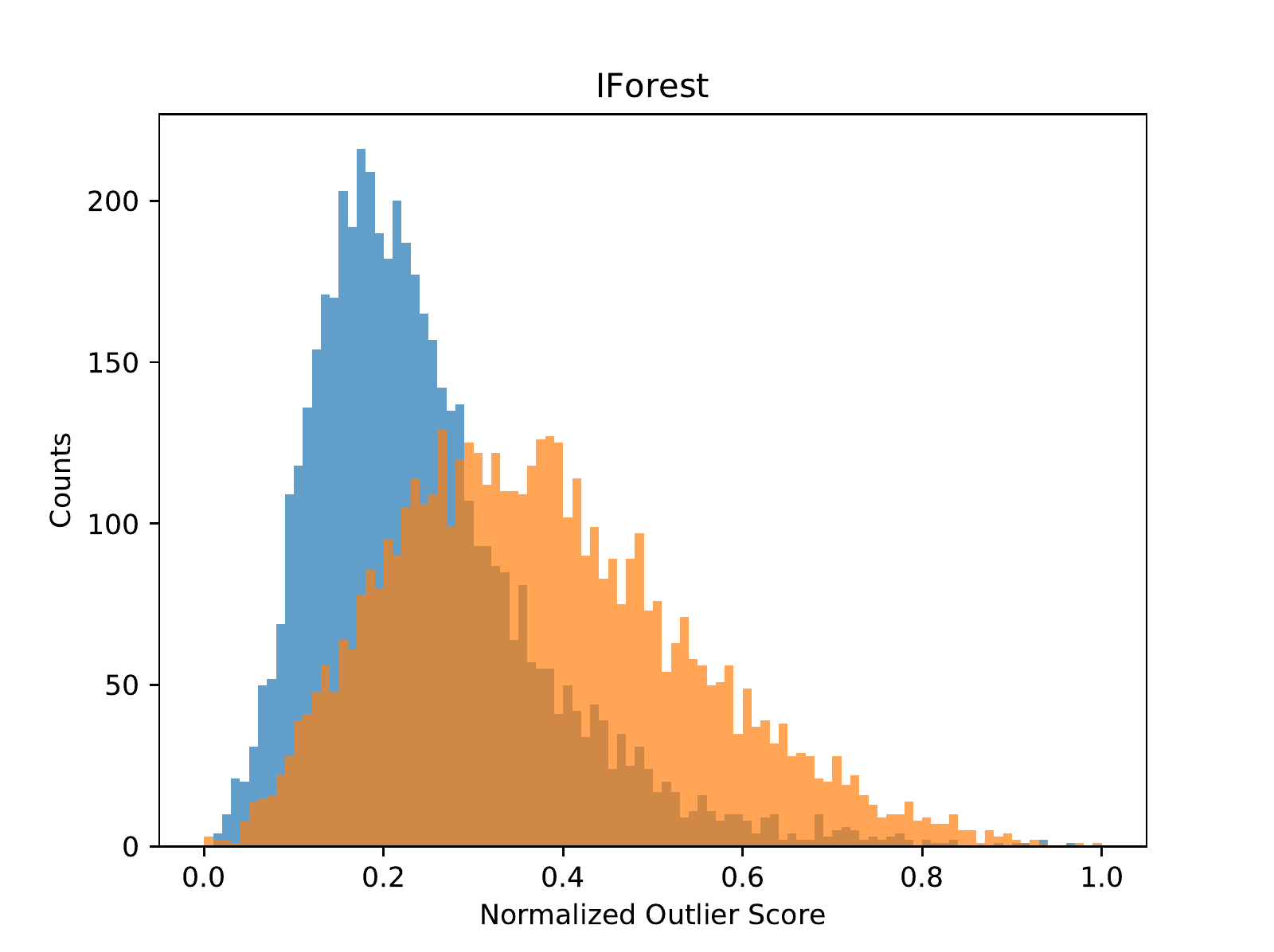}
                \caption{IForest \cite{IForest}}
                \label{fig:IForest_Hist}
            \end{subfigure}
            \begin{subfigure}[t]{.24\linewidth}
                \centering
                \includegraphics[width=\textwidth]{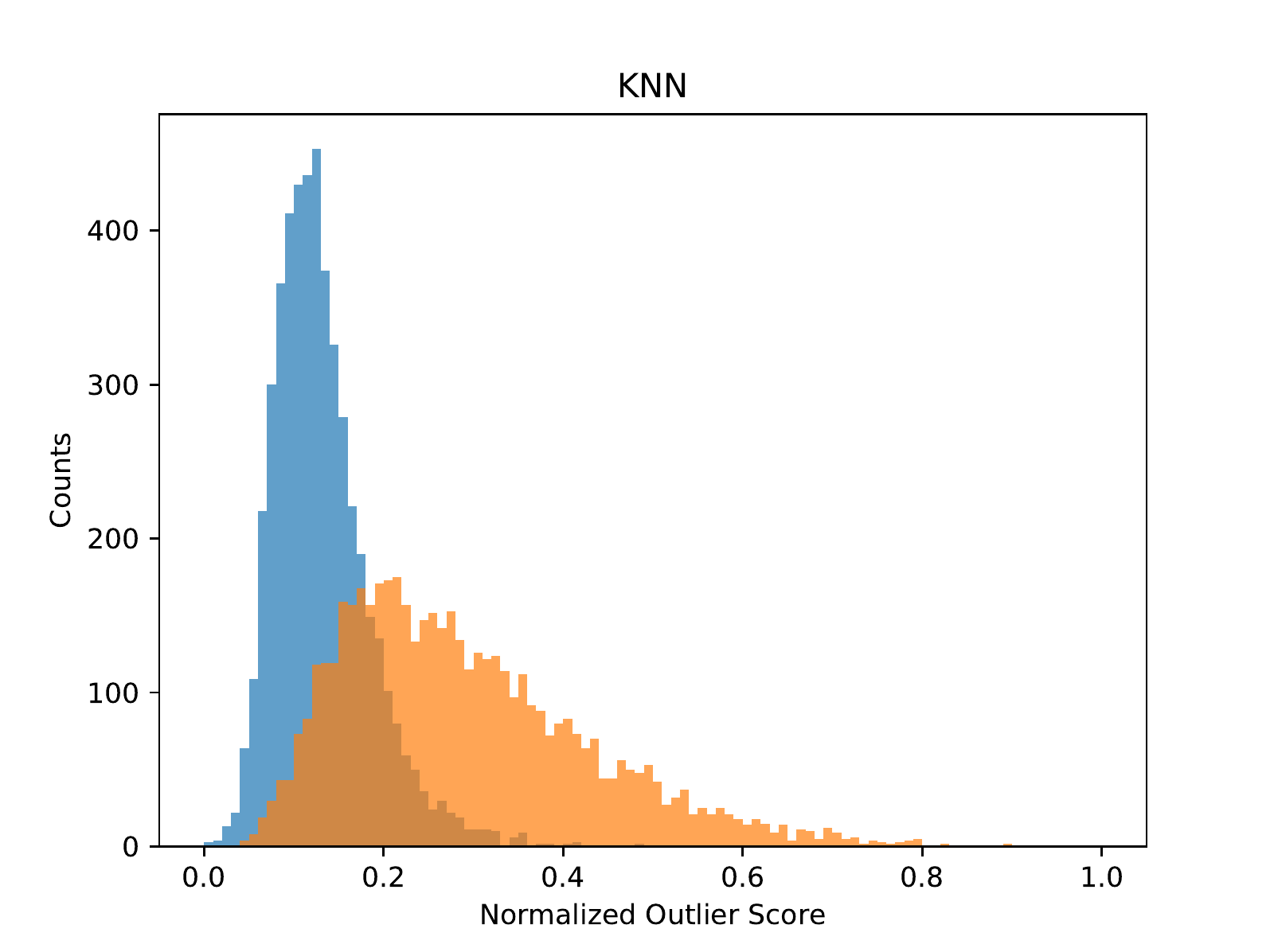}
                \caption{KNN \cite{KNN}}
                \label{fig:KNN_Hist}
            \end{subfigure}\\
            \begin{subfigure}[t]{.24\linewidth}
                \centering
                \includegraphics[width=\textwidth]{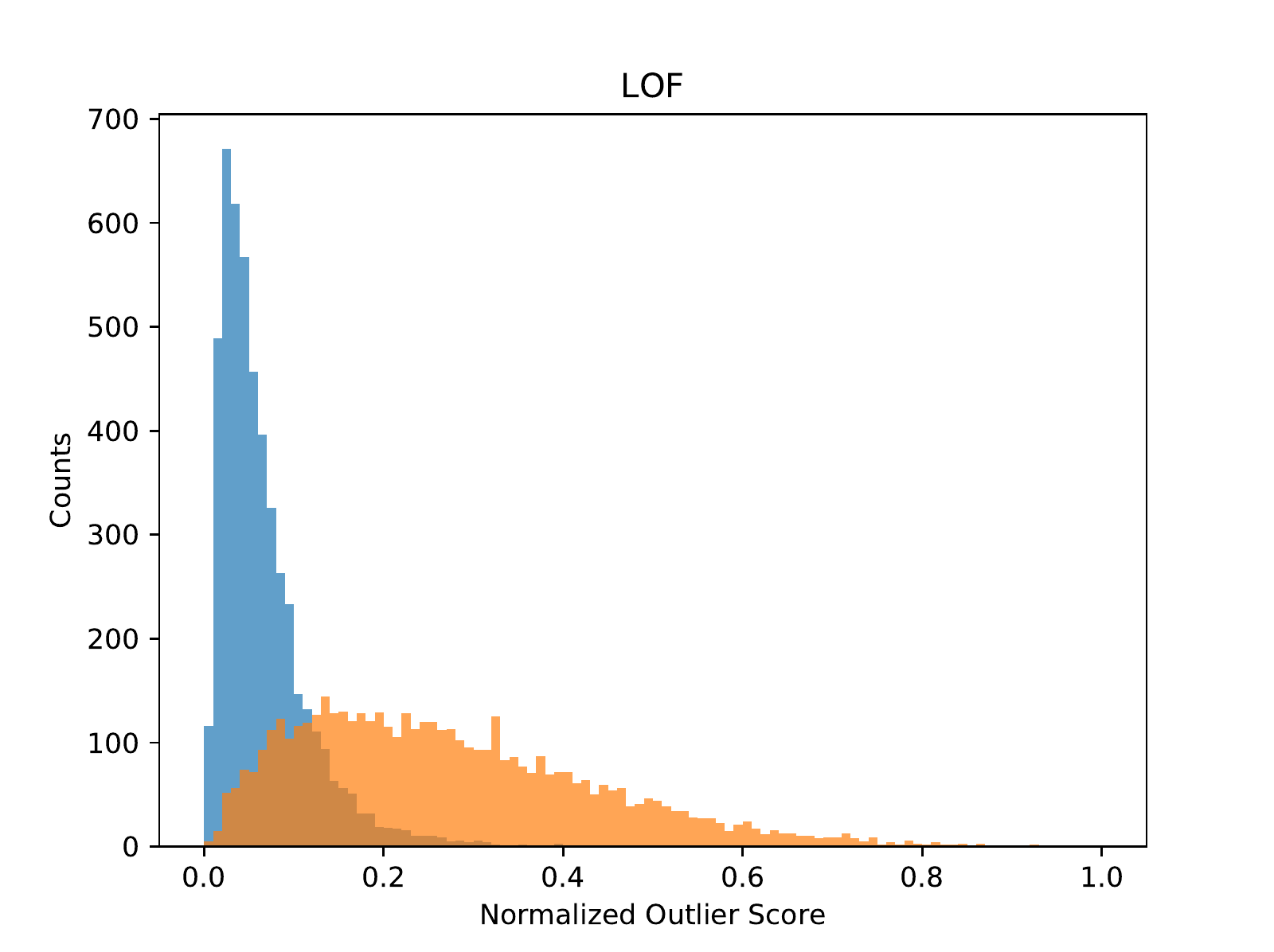}
                \caption{LOF \cite{LOF}}
                \label{fig:LOF_Hist}
            \end{subfigure}
            \begin{subfigure}[t]{.24\linewidth}
                \centering
                \includegraphics[width=\textwidth]{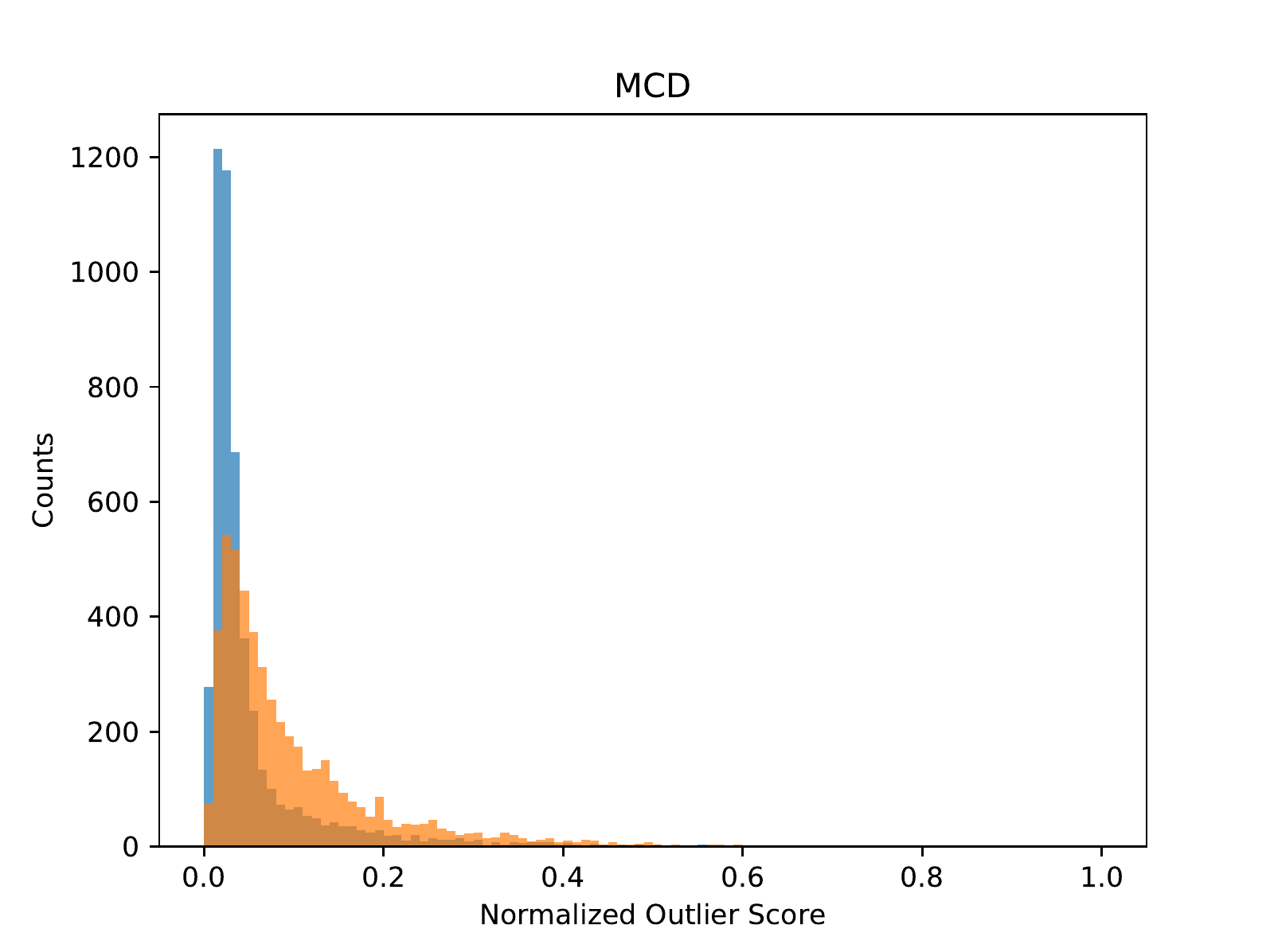}
                \caption{MCD \cite{MCD}}
                \label{fig:MCD_Hist}
            \end{subfigure}
            \begin{subfigure}[t]{.24\linewidth}
                \centering
                \includegraphics[width=\textwidth]{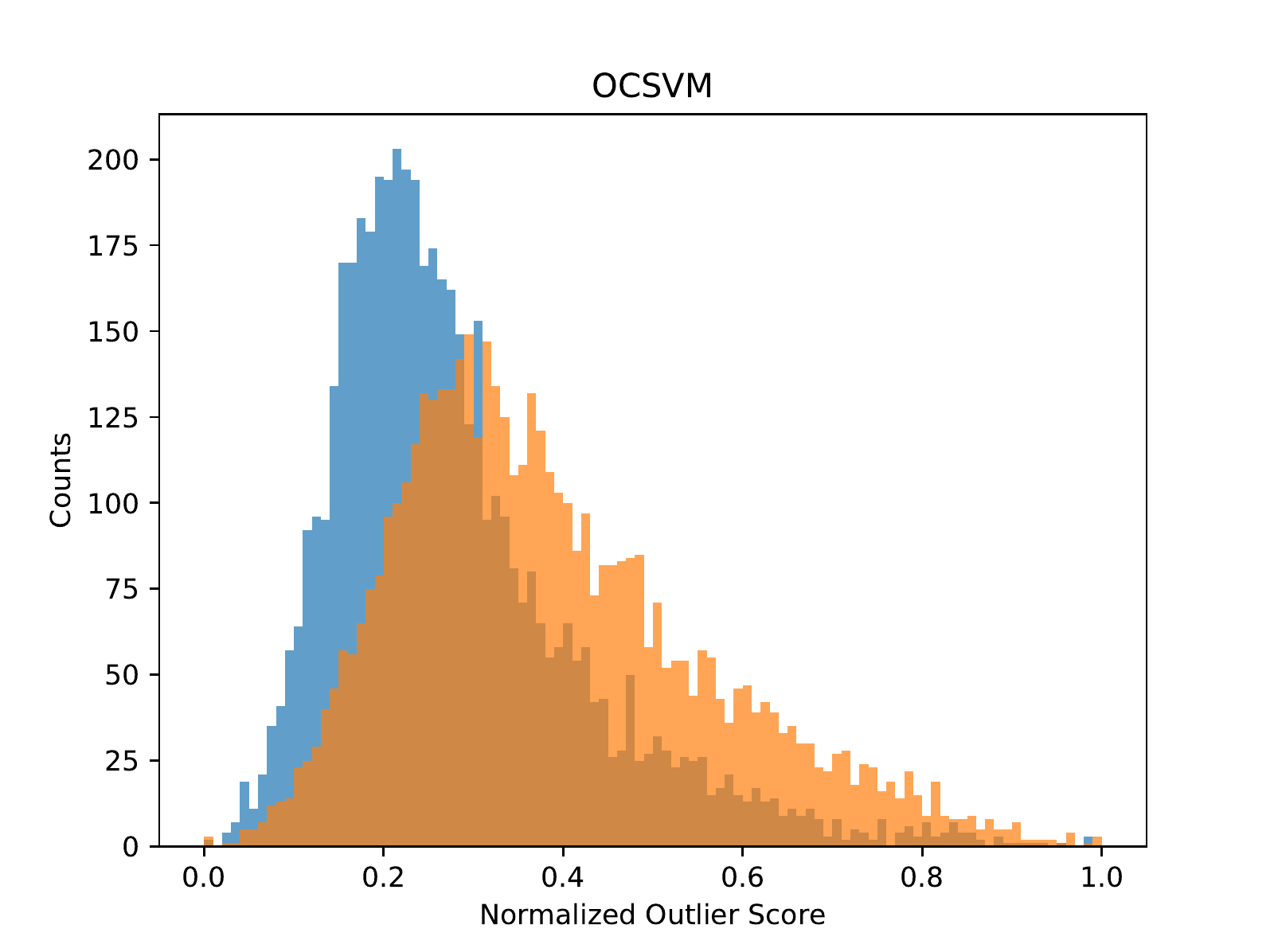}
                \caption{OCSVM \cite{OCSVM}}
                \label{fig:OCSVM_Hist}
            \end{subfigure}
            \begin{subfigure}[t]{.24\linewidth}
                \centering
                \includegraphics[width=\textwidth]{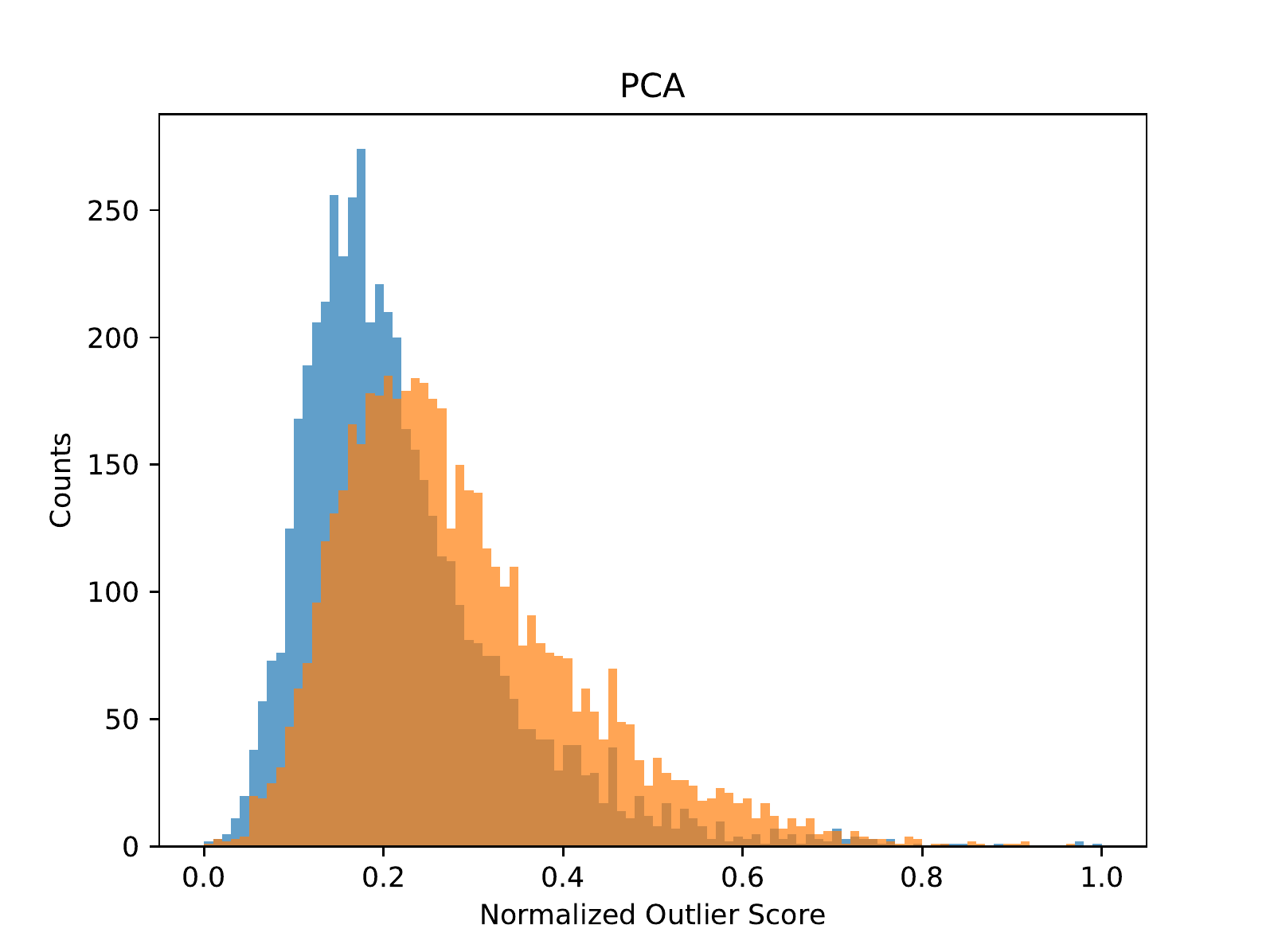}
                \caption{PCA \cite{PCA}}
                \label{fig:PCA_Hist}
            \end{subfigure}\\
            \begin{subfigure}[t]{.24\linewidth}
                \centering
                \includegraphics[width=\textwidth]{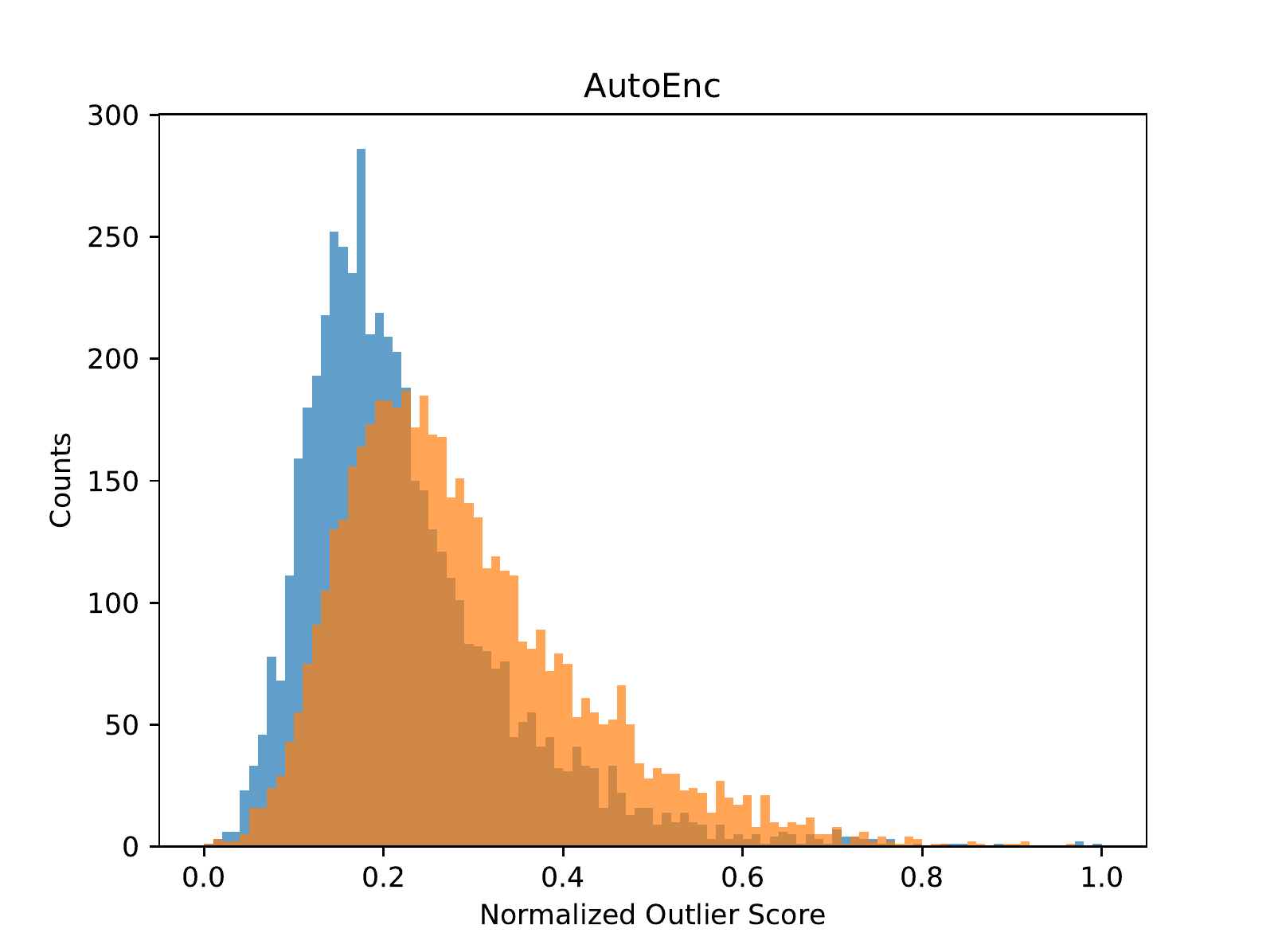}
                \caption{AutoEnc \cite{AutoEnc}}
                \label{fig:AutoEnc_Hist}
            \end{subfigure}
            \begin{subfigure}[t]{.24\linewidth}
                \centering
                \includegraphics[width=\textwidth]{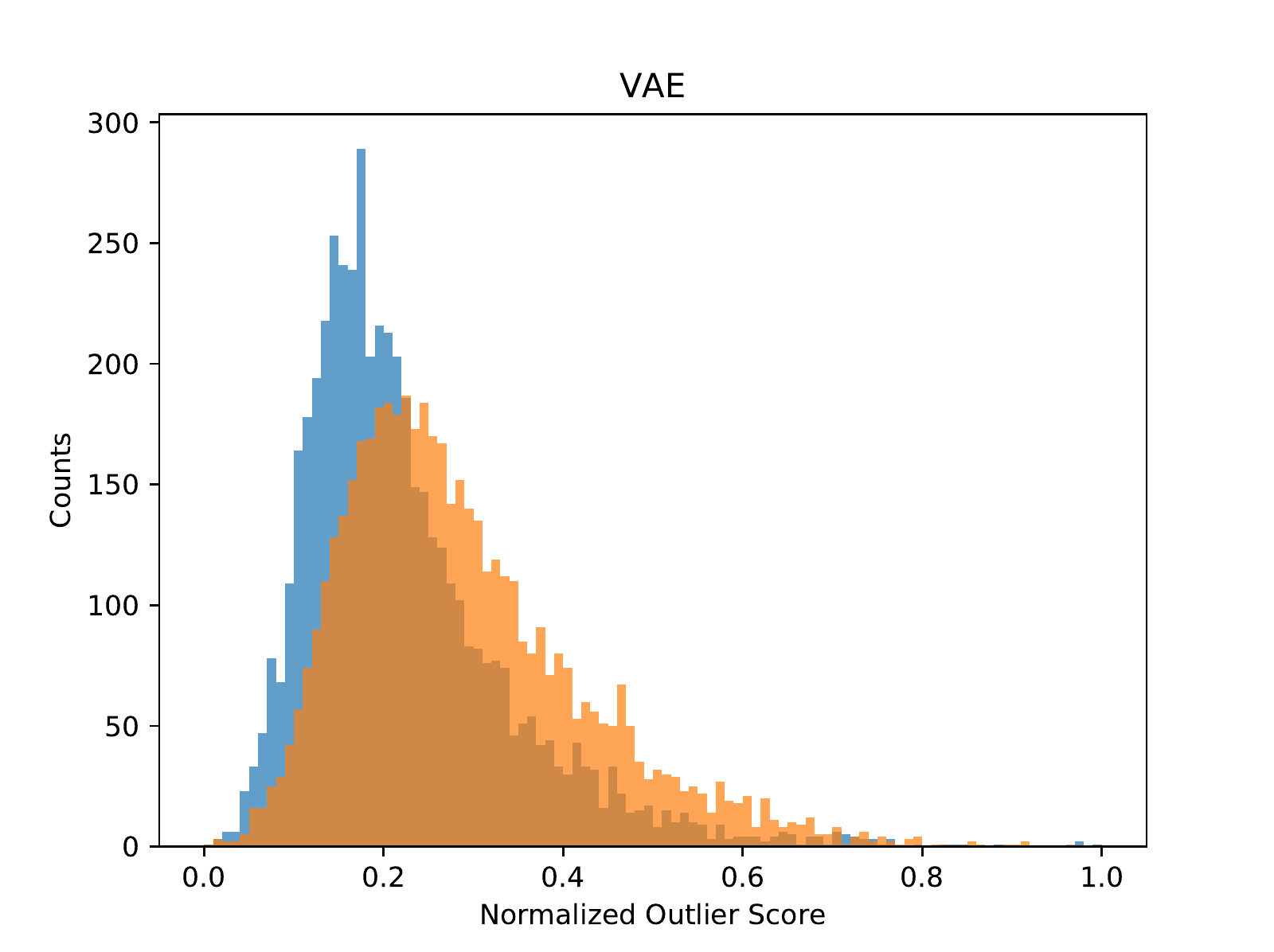}
                \caption{VAE \cite{VAE}}
                \label{fig:VAE_Hist}
            \end{subfigure}
            \begin{subfigure}[t]{.24\linewidth}
                \centering
                \includegraphics[width=\textwidth]{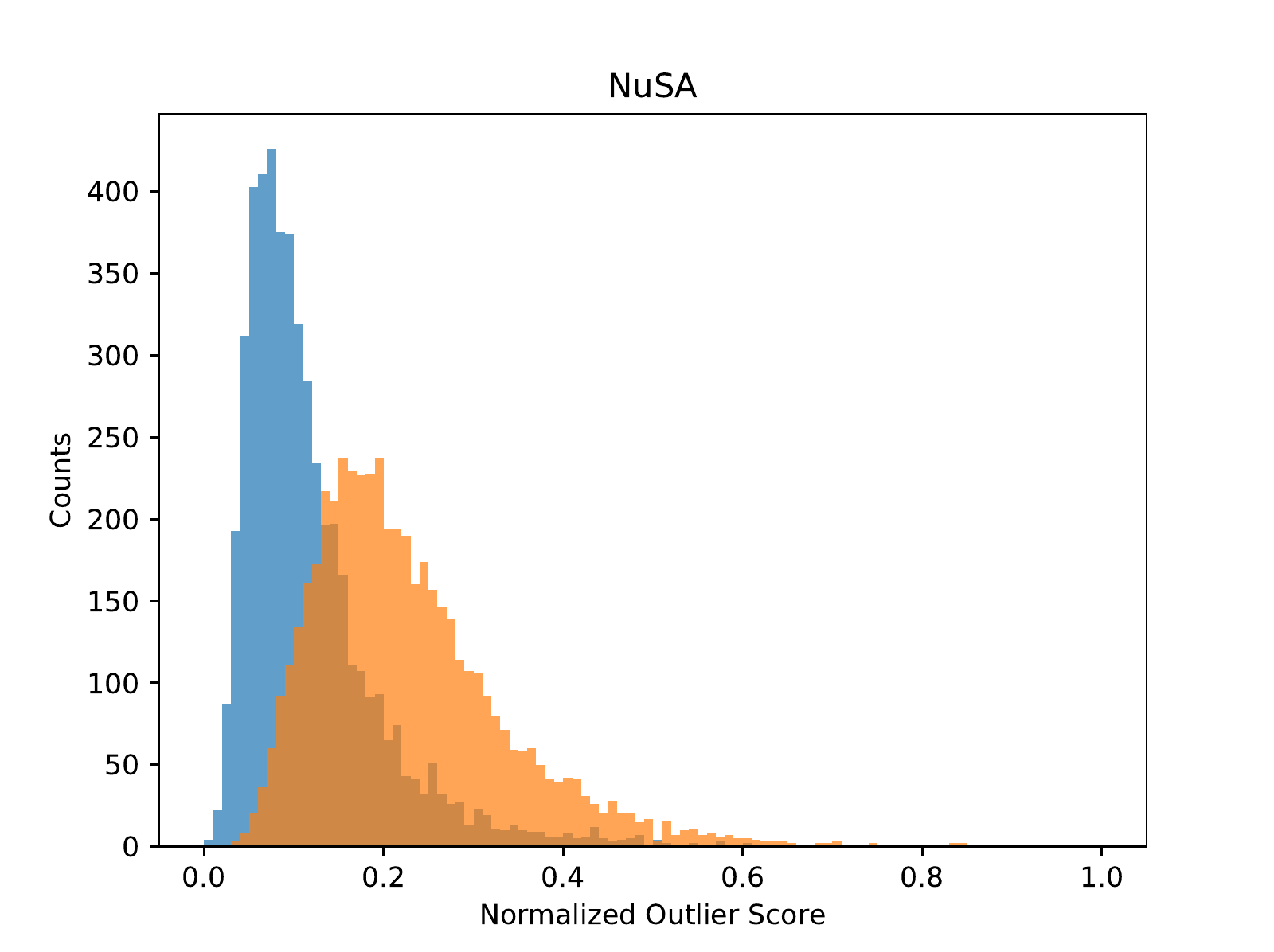}
                \caption{NuSA}
                \label{fig:NuSA_Hist}
            \end{subfigure}
            \caption{Histograms of each algorithms outlier detection scores. In all images the blue histogram represents the data from the known classes (0, 2, 5, 7, and 9 in this case) and the orange represents the unknown classes. The output from NuSA has been inverted in this figure so that as the score increases the likelihood of being an outlier increases to match the other algorithms. The outputs from each of the algorithms have been scaled to fall between zero and one.}
            \label{fig:Hists}
        \end{figure*}

        In Figure \ref{fig:Hists} we show a particular set of results. In the results 
        shown we highlight the case where the known classes are 0, 2, 5, 7, and 9. For 
        each method we make a histogram of the algorithms' output values for known and 
        unknown classes to investigate the separation between the two classes. In all 
        cases the orange histogram shows the known classes while the blue shows the 
        unknown classes. In these histograms we can see that NuSA generates
        two overlapping distributions for the known and unknown classes. The 
        other algorithms that perform the best are ABOD, KNN, and LOF these methods
        outperform NuSA here due to the compactness of their distributions. Yet, NuSA 
        appears to actually have a larger separation between the means of the 
        distributions.
        
    {\small
        \bibliographystyle{icml2020}
        \bibliography{NS_Paper}
    }

\end{document}